\documentclass[12pt]{article}
\usepackage{graphicx}
\usepackage{enumerate}
\usepackage{natbib}
\usepackage{url} 

\newcommand{\blind}{1}

\usepackage{amsmath}
\usepackage{amsfonts}
\usepackage{mathtools}
\usepackage{mathrsfs}
\numberwithin{equation}{section}
\usepackage{amssymb}
\usepackage{multicol}
\usepackage{mathtools}
\usepackage{tikz}
\usepackage{setspace}
\usepackage{booktabs}
\usepackage{mathtools}
\usepackage{array}
\usepackage{graphicx}
\usepackage{xcolor}
\usepackage{float}
\usepackage[colorlinks=true, 
    citecolor={blue}]{hyperref}
\usepackage{amsthm}
\usepackage{enumitem}
\usepackage{subcaption}
\usepackage{pgfplots}
\pgfplotsset{compat=1.18}
\usetikzlibrary{positioning, shapes}
\usetikzlibrary{matrix}
\usetikzlibrary{shapes.symbols}

\usepackage{xr}
\definecolor{madridblue}{RGB}{0, 102, 204}
\definecolor{lightblue1}{HTML}{0748CD}
\definecolor{lightblue2}{HTML}{3163EB}
\definecolor{lightblue3}{HTML}{5882F8}
\definecolor{lightblue4}{HTML}{84A1F9}
\definecolor{lightblue5}{HTML}{ADBFFB}

\definecolor{lightgreen1}{HTML}{8CFEBA}
\definecolor{lightgreen2}{HTML}{A8FEAB}
\definecolor{lightgreen3}{HTML}{C4FE9D}
\definecolor{lightgreen4}{HTML}{E0FD8E}
\definecolor{lightgreen5}{HTML}{FCFC7F}

\definecolor{green1}{HTML}{13533E}
\definecolor{green2}{HTML}{23965D}
\definecolor{green3}{HTML}{43BB73}
\definecolor{green4}{HTML}{81D5A2}
\definecolor{green5}{HTML}{C1E7CD}

\definecolor{lightpurple1}{HTML}{8834B3}
\definecolor{lightpurple2}{HTML}{AE4AD9}
\definecolor{lightpurple3}{HTML}{D664F8}
\definecolor{lightpurple4}{HTML}{E48CF9}
\definecolor{lightpurple5}{HTML}{F0B5FF}

\definecolor{lightred1}{HTML}{D90070}
\definecolor{lightred2}{HTML}{FF007D}
\definecolor{lightred3}{HTML}{FF529A}
\definecolor{lightred4}{HTML}{FF7AAD}
\definecolor{lightred5}{HTML}{FFA3C3}

\definecolor{red1}{HTML}{941B14}
\definecolor{red2}{HTML}{D0241C}
\definecolor{red3}{HTML}{F54D40}
\definecolor{red4}{HTML}{F9877D}
\definecolor{red5}{HTML}{F7B3AC}

\definecolor{orange1}{HTML}{91341D}
\definecolor{orange2}{HTML}{C65323}
\definecolor{orange3}{HTML}{E96B3D}
\definecolor{orange4}{HTML}{F9885E}
\definecolor{orange5}{HTML}{FAB590}

\definecolor{uciblue}{RGB}{0,100,164}
\definecolor{ucigray}{RGB}{197,190,181}
\definecolor{ucigold}{RGB}{255,210,0}
\definecolor{ucilightblue}{RGB}{106,162,184}
\definecolor{ucidarkblue}{RGB}{27,61,109}
\definecolor{ucilimegreen}{RGB}{122,184,0}
\definecolor{uciorange}{RGB}{247,141,45}

\newtheorem{assumption}{Assumption}[section]

\newtheorem{lemma}{Lemma}[section]
\newtheorem{remark}{Remark}[section]

\newtheorem{theorem}{Theorem}[section]
\newtheorem{corollary}{Corollary}[section]

\DeclareMathOperator*{\argmin}{arg\,min}

\begin{document}

\def\spacingset#1{\renewcommand{\baselinestretch}%
{#1}\small\normalsize} \spacingset{1}


\if1\blind
{
  \title{\bf Representation Retrieval Learning for Heterogeneous Data Integration}
  \author{Qi Xu \\
    Department of Statistics \& Data Science,\\ Carnegie Mellon University\\
    and \\
    Annie Qu\thanks{Corresponding author. The authors gratefully acknowledge \textit{NSF Grant DMS 2210640 and NCI R01 grant 1R01CA297869-01.}}\hspace{.2cm} \\
    Department of Statistics and Applied Probability,\\ University of California, Santa Barbara}
  \maketitle
} \fi

\if0\blind
{
  \bigskip
  \bigskip
  \bigskip
  \begin{center}
    {\LARGE\bf Representation Retrieval Learning for Heterogeneous Data Integration}
\end{center}
  \medskip
} \fi

\bigskip
\begin{abstract}
    In the era of big data, large-scale, multi-source, multi-modality datasets are increasingly ubiquitous, offering unprecedented opportunities for predictive modeling and scientific discovery. However, these datasets often exhibit complex heterogeneity, such as covariates shift, posterior drift, and blockwise missingness, which worsen predictive performance of existing supervised learning algorithms. To address these challenges simultaneously, we propose a novel Representation Retrieval ($R^2$) framework, which integrates a dictionary of representation learning modules (\emph{representer dictionary}) with data source-specific sparsity-induced machine learning model (\emph{learners}). Under the $R^2$ framework, we introduce the notion of \emph{integrativeness} for each representer, and propose a novel \emph{Selective Integration Penalty} (SIP) to explicitly encourage more integrative representers to improve predictive performance. Theoretically, we show that the excess risk bound of the $R^2$ framework is characterized by the \emph{integrativeness} of representers, and SIP effectively improves the excess risk. Extensive simulation studies validate the superior performance of $R^2$ framework and the effect of SIP. We further apply our method to two real-world datasets to confirm its empirical success.
\end{abstract}

\noindent%
\vfill

\newpage
\spacingset{1.9} 
\section{Introduction}
\label{sec: intro}

Large-scale data integration has made transformative contributions across numerous fields, including computer vision, natural language processing, biomedicine, genomics and healthcare. For example, in biomedicine, integrating randomized clinical trials and observational studies is of great interest, as it borrows information from both data sources \citep{yang2020combining, shi2023data, colnet2024causal}. In genomics, multi-modality and multi-batch assays enable the discovery of cellular heterogeneity and development \citep{kriebel2022uinmf, du2022robust}. In healthcare, multiple types of time-series measurements, such as cardiovascular, physical activities, and sleep data can be integrated to improve real-time health and well-being monitoring \citep{zhang2024individualized, sun2025generalized, rim2025individualized}. However, integration of large-scale data effectively remains challenging, particularly when data are collected from diverse sources or populations, and across various variables and modalities.

Integrating large-scale data is particularly challenging due to various types of heterogeneity. First, the marginal distributions of the same modality are often heterogeneous across different sources or populations, a phenomenon called ``distribution heterogeneity'', or ``covariates shift'' in the literature \citep{shimodaira2000improving}. Second, in the context of supervised learning, the conditional distribution of responses given covariates could be heterogeneous, which is named ``posterior heterogeneity'', or ``posterior drift'' \citep{widmer1996learning}. Third, observed covariates or modalities are often not uniformly measured: some covariates are observed across all data sources, while others are observed in only partial data sources. We refer to this as ``observation heterogeneity'' or ``blockwise missingness'' \citep{yu2020optimal, xue2021integrating}. 

In this work, we propose a unified framework to incorporate all three types of heterogeneity in large-scale data integration. Specifically, we introduce the Representation Retrieval ($R^2$) framework, which constructs a dictionary of representation learning modules (\emph{representer}), such as neural nets \citep{goodfellow2016deep} and smoothing function bases \citep{ravikumar2009sparse}, to capture the complex distribution across all data sources. Then, each data source is allowed to retrieve relevant and informative representers from the dictionary for prediction to accommodate covariates shift among data sources. Finally, source-specific machine learning algorithms (\emph{learner}) are employed to predict responses based on the retrieved latent representations, such that the posterior heterogeneity is incorporated. This learning algorithm is named as Representation Retrieval ($R^2$) learning hereafter. Further, we extend $R^2$ learning to Blockwise Representation Retrieval ($BR^2$) learning to handle the observation heterogeneity. In particular, we introduce modality-specific representer dictionaries and aggregate retrieved representer from each modality for prediction. This approach exploits all available observed data and does not require missing imputation, which is robust against untestable missing mechanism assumptions.

Moreover, we introduce a new concept of       \emph{integrativeness} of representers, defined as the number of data sources retrieving the specific representer. Theoretically, we provide a precise excess risk bound of the $R^2$ framework, and reveals that the upper bound is explicitly controlled by \emph{integrativeness} of represeneters. Motivated by this phenomenon, we propose a novel Selective Integration Penalty (SIP), and explicitly encourage more integrative representers to improve prediction performance. Extensive simulation studies and real-world applications also confirm the superior empirical performance of our proposed method.

\subsection{Related works}

In the current literature, various approaches for handling heterogeneous data have been proposed, while most of them only address one or two types of the aforementioned heterogeneity. We review the related works as follows.

Covariates shift is encountered frequently in data integration, and has been studied extensively in the literature. Related methods can be summarized into two categories: first, joint-and-individual structure, assuming a joint component that shared by all data sources, while each data source has its unique component to capture the individual information not explained by the joint component. Some representative works include \citep{lock2013joint, klami2014group, bunte2016sparse, feng2018angle, tang2021integrated, lin2023quantifying, wang2024joint}. Second, partially sharing structure  \citep{gaynanova2019structural, lock2022bidimensional, xiao2024sparse} is a more flexible framework that each component can be shared by any numbers of data sources, to accommodate greater heterogeneity. In the latent space modeling, a shared latent space is assumed and each data source has different projection onto this shared latent space \citep{klami2014group, bunte2016sparse, tang2021integrated}, which fits into the joint-and-individual structure. In contrast, our method assumes a shared dictionary and allows each data source to retrieve any representers from the dictionary, following the partially sharing structure. Therefore, our method is more adaptive to greater heterogeneity in covariates shift. Instead of uncovering the partially sharing structure in unsupervised learning as in \citep{gaynanova2019structural, lock2022bidimensional, xiao2024sparse}, our proposed method actively encourages partially sharing structure via SIP to improve predictive performance.

Posterior heterogeneity has been extensively studied in the literature of transfer learning and multi-task learning. There are various structural assumptions imposed on model parameters. For instance, regression coefficients across tasks are assumed to be similar, and can be achieved by a fusion-type penalty \citep{tang2016fused}, or shrinking model parameters to a prototype \citep{duan2023adaptive}. Other works consider more complex structures, such as latent representation \citep{tian2023learning}, or angular similarity \citep{gu2024robust}. In these parametric models, parameters are stringent to be comparable across data sources, implying that models for all sources belong to the same function class. In contrast, our method can include representers of varying complexity and each data source can retrieve informative ones to maximize their prediction accuracy.

Observation heterogeneity, or blockwise missingness often has been treated separately from the other two heterogeneity. In multi-view learning with missing views \citep{xu2015multi, zhang2018multi, fan2023incomplete, xie2023exploring}, the primary target is to impute missing views by leveraging some structural assumptions, such as low-rank structure \citep{cai2016structured, zhou2023multi}. Other works target prediction with the presence of missing modalities \citep{yuan2012multi, yu2020optimal, xue2021integrating, yu2022integrative, song2024semi, sell2024nonparametric, sui2025multi, ma2025deep}. These methods can be summarized into imputation \citep{yuan2012multi, xue2021integrating, sui2025multi, ma2025deep} and imputation-free \citep{yu2020optimal, yu2022integrative, song2024semi, sell2024nonparametric} methods. In general, imputation approaches impose more stringent requirement on missing mechanisms than imputation-free approach. When the assumed missing mechanism is violated, or imputations are biased, imputation approaches could suffer from degraded performance. Our method is imputation-free that utilizes all observed data, and robust against missing mechanism misspecification.

\subsection{Organization}

The rest of paper is structured as follows: In Section \ref{sec: prelim}, we introduce the notations used throughout the paper and elaborate the problem setup in this work. In Section \ref{subsec: R2learn}, we first present the Representation Retrieval ($R^2$) learning for multi-task learning problem. Then, we present the notion of \emph{integrativeness} and introduce the Selective Ingration Penalty (SIP) in Section \ref{subsec: sip}. The extension to Blockwise Representation Retrieval ($BR^2$) learning is elaborated in Section \ref{subsec: BR2learn}. Theoretical analyses of the $R^2$ framework are provided in Section \ref{sec: theory}, which recalls the important role of SIP in improving predictive performance. Afterwards, we investigate the empirical performance of our $R^2$ and $BR^2$ learning with simulated data in Section \ref{sec: simulation}, and apply our method to two real data examples to demonstrate its superior performance in Section \ref{sec: real-data}. Summary and discussion are provided in Section \ref{sec: discussion}.

\section{Preliminaries}
\label{sec: prelim}

In this section, we introduce our problem setup with essential notations. The related background is also introduced as a prelude to the proposed method.

In this paper, we focus on integrating multi-source and multi-modality data for supervised learning problems, such as classification and regression. Suppose we collect $S$ data sources and $M$ modalities: $\mathcal{D}_{s} = (\mathbf{X}^{(s)}, \mathbf{y}^{(s)})$, where $\mathbf{X}^{(s)} \in \mathcal{X}^{(s)} \subset \mathbb{R}^{p_s}$ denotes observed modalities, and $\mathbf{y}^{(s)} \in \mathcal{Y}^{(s)}$ denotes responses or outcomes. Note that $\mathcal{X}^{(s)}$ lies in $\mathbb{R}^{p_s}$ where the dimensionality $p_s$ depends on the observing modalities of the $s$th data source. More detailed notations for multi-modality covariates will be introduced in Section \ref{subsec: BR2learn}. The response spaces $\mathcal{Y}^{(s)}$'s are source-specific to accommodate heterogeneous responses, for example, $\mathcal{Y}^{(s)} = \{1, 2, ..., \}$ for classification problems and $\mathcal{Y}^{(s)} \in \mathbb{R}$ for regression problems. A straightforward illustration of our setup is provided in the upper block of Figure \ref{fig: special_cases}. As noted in Introduction, our problem setup encompasses a broad spectrum of problem setups, including multi-task learning and blockwise missing learning as special cases. Our target is to learn function mappings $f^{(s)}: \mathcal{X}^{(s)} \rightarrow \mathcal{Y}^{(s)}$ for $s = 1, \cdots, S$. Although it is valid to learn $f^{(s)}$ from the $s$th data source solely, here we seek to integrate shared information across different data sources and modalities to strengthen predictive performance for each source.

\begin{figure}[t]
    \centering
    \scalebox{0.55}{
    \begin{tikzpicture}
    \node (placeholder1) at (-5, 2) {}; 
    \node (placeholder2) at (1, 2) {};
    \node (placeholder3) at (-5, -2) {};
    \draw[->, very thick] (placeholder1) -- (placeholder2) node[midway, above] {Modalities};
    \draw[->, very thick] (placeholder1) -- (placeholder3) node[midway, sloped, above, rotate=180] {Sources};
    
    \matrix[matrix of nodes, ampersand replacement=\&, nodes={draw, minimum width=10mm, minimum height=5mm, anchor=center}] (data) at (-2, 0) {
    |[fill=lightblue1]|\& |[fill=green3]| \& |[fill=white]| \& |[fill=lightred2]|\& |[fill=red3]| \\
    |[fill=lightblue4]| \& |[fill=white]| \& |[fill=lightpurple3]| \& |[fill=lightred4]| \& |[fill=red5]| \\
    |[fill=white]| \& |[fill=green2]| \& |[fill=lightpurple2]| \& |[fill=lightred3]| \& |[fill=white]| \\
    |[fill=lightblue3]| \& |[fill=green5]| \& |[fill=white]| \& |[fill=white]| \& |[fill=red4]| \\
    |[fill=lightblue2]| \& |[fill=white]| \& |[fill=lightpurple1]| \& |[fill=lightred1]| \& |[fill=red3]| \\};

    \node (X_label) at (-2, -2) {Covariates};

    \matrix[matrix of nodes, ampersand replacement=\&, nodes={draw, minimum width=10mm, minimum height=5mm, anchor=center}] (response) at (3, 0) {
    |[fill=uciblue]| \\
    |[fill=ucigold]| \\
    |[fill=ucigray]| \\
    |[fill=ucilightblue]| \\
    |[fill=uciorange]| \\
    };

    \node (y_label) at (3, -2) {Responses};

    \node [rounded corners, thick, draw=black, fill=blue!20, minimum width = 12cm, minimum height = 6cm, fill opacity = 0.3] (proposal) at (0, 0) {};

    \node [rounded corners, thick, draw=black, fill=green!20,
    minimum width = 8cm, minimum height = 4cm, fill opacity = 0.3] (MTL) at (-6, -6) {};

    \draw[->, very thick] (proposal) -- (MTL);

    \matrix[matrix of nodes, ampersand replacement=\&, nodes={draw, minimum width=4cm, minimum height=4mm, anchor=center}] (data) at (-7, -6) {
    |[fill=lightblue1]| \\
    |[fill=lightblue4]| \\
    |[fill=lightblue1]| \\
    |[fill=lightblue3]| \\
    |[fill=lightblue2]| \\};

    \node (X_label2) at (-7, -7.5) {Covariates};

    \matrix[matrix of nodes, ampersand replacement=\&, nodes={draw, minimum width=8mm, minimum height=4mm, anchor=center}] (response) at (-3, -6) {
    |[fill=uciblue]| \\
    |[fill=ucigold]| \\
    |[fill=ucigray]| \\
    |[fill=ucilightblue]| \\
    |[fill=uciorange]| \\
    };

    \node (y_label2) at (-3, -7.5) {Responses};
    \node (MTL_label) at (-6, -4.5) {Multi-task learning};







    \node [rounded corners, thick, draw=black, fill=orange!20,
    minimum width = 8cm, minimum height = 4cm, fill opacity = 0.3] (blockmissing) at (6, -6) {};

    \draw[->, very thick] (proposal) -- (blockmissing);

    \matrix[matrix of nodes, ampersand replacement=\&, nodes={draw, minimum width=8mm, minimum height=4mm, anchor=center}] (data) at (5, -6) {
    |[fill=lightblue5]|\& |[fill=green4]| \& |[fill=white]| \& |[fill=lightred4]|\& |[fill=red5]| \\
    |[fill=lightblue5]| \& |[fill=white]| \& |[fill=lightpurple5]| \& |[fill=lightred4]| \& |[fill=red5]| \\
    |[fill=white]| \& |[fill=green4]| \& |[fill=lightpurple5]| \& |[fill=lightred4]| \& |[fill=white]| \\
    |[fill=lightblue5]| \& |[fill=green4]| \& |[fill=white]| \& |[fill=white]| \& |[fill=red5]| \\
    |[fill=lightblue5]| \& |[fill=white]| \& |[fill=lightpurple5]| \& |[fill=lightred4]| \& |[fill=red5]| \\};

    \matrix[matrix of nodes, ampersand replacement=\&, nodes={draw, minimum width=8mm, minimum height=4mm, anchor=center}] (response) at (9, -6) {
    |[fill=ucigold]| \\
    |[fill=ucigold]| \\
    |[fill=ucigold]| \\
    |[fill=ucigold]| \\
    |[fill=ucigold]| \\
    };

    \node (blockmissing_label) at (6, -4.5) {Block-wise missing learning};

    \node (X_label3) at (5, -7.5) {Covariates};
    \node (y_label3) at (9, -7.5) {Responses};
\end{tikzpicture}
    }
    \caption{\small Targeted problem setup and special cases. The upper block illustrates our problem setup: different colors represent different covariates modalities, and white blocks denote missing modalities. Different shade levels of the same color indicate different distributions of the same covariates. The lower left block shows the multi-task learning setup, which considers only one covariate modality and ignores observation heterogeneity. The lower right block illustrates block-wise missing learning, involving multi-modality data with missing blocks without considering distribution and posterior heterogeneity.}
    \label{fig: special_cases}
\end{figure}

\subsection{Multi-task representation learning}
\label{subsec: rl_mtl}

In the machine learning community, representation learning has received considerable attention due to its powerful and flexible capacity to learn complex and non-linear patterns. A comprehensive review can be found in \citep{bengio2013representation}. For supervised learning settings, 
a typical procedure following representation learning is: $\mathbf{X} \underbrace{\stackrel{\theta}{\longrightarrow} \mathbf{h} \stackrel{g}{\longrightarrow} }_{f} \mathbf{y}$, where $\theta$ is a latent representation algorithm (\textit{representer}) which projects the original data into a latent space, and $g$ is a regression or classification algorithm (\textit{learner}) which establishes the association between latent representation and responses. In other words, the function $f$ is a composition of $g$ and $\theta$, denoted by $f = g\circ \theta$. This learning paradigm is a building block in deep learning for many applications. The \textit{representer} $\theta$ is usually a complex neural net to uncover the nonlinear, entangled features, which allows a simple \textit{learner} $g$ to achieve desirable empirical performance. In multi-task representation learning \citep{du2020few, tripuraneni2021provable, watkins2023optimistic}, each task is associated with a unique $g^{(s)}$, so the task-specific function mapping is formulated as $f^{(s)} = g^{(s)} \circ \theta$. In words, multi-task representation learning utilizes $\theta$ to learn shared representation across all tasks, and employs $g^{(s)}$ to capture all heterogeneity among tasks. In this framework, source-specific $g^{(s)}$'s are designed to capture all heterogeneity among data sources, however, it usually fail to accommodate both covariates shift and posterior drift. In this work, we employ representation learning to capture complex and non-linear patterns, and propose a new framework to address the above drawback in multi-task representation learning.

\section{Methodology}
\label{sec: method}

\subsection{Representation Retrieval Learning}
\label{subsec: R2learn}

We first consider integrating multi-source datasets with the same modalities for supervised learning. This is equivalent to the multi-task learning problem with both covariates shift and posterior drift. Therefore, we use the terms ``tasks'' and ``sources'' interchangeably in this subsection.

We consider a representer dictionary as a collection of representers, denoted as $\Theta = \{\theta_1, \cdots, \theta_D\}$, where $D$ is the number of representers contained in $\Theta$. Formally, we define a representer as a univariate function mapping $\theta_d: \mathcal{X} \rightarrow \mathbb{R}$, so we use $\theta_d$ and $\theta_d(\mathbf{x})$ interchangeably in later discussion. Multivariate representer outputs are also allowed, where the related extension can be found in Appendix \ref{app_sec: multi-dim-rep}. To accommodate covariates shift among tasks, we assume that a subset of representers from $\Theta$ are informative to predict $y^{(s)}$ for the $s$th task, denoted as $\Gamma^{(s)} = \{\theta_{d_1}, \cdots, \theta_{d_s}\}$. Naturally, we define $\Gamma^{(s)}(\mathbf{x}) = (\theta_{d_1}(\mathbf{x}), \cdots, \theta_{d_s}(\mathbf{x})): \mathcal{X} \rightarrow \mathbb{R}^{|\Gamma^{(s)}|}$. Since we do not impose any structural assumptions on $\Gamma^{(s)}$, both joint-and-individual \citep{lock2013joint} and partially sharing \citep{gaynanova2019structural} structures can be incorporated.

In order to learn the function mappings $f^{(s)}$ and account for posterior drift, we build the task-specific learners upon the retrieved representers as in multi-task representation learning. Suppose $g^{(s)}(\cdot) = \langle\cdot, \alpha^{(s)}\rangle$ is a linear function, then each function mapping $f^{(s)}$ is formulated as 
{\small
\begin{align}
    \label{f_formula1}
    f^{(s)}(\cdot) = \langle \Gamma^{(s)}(\cdot), \alpha^{(s)}\rangle, \quad \alpha^{(s)} \in \mathbb{R}^{|\Gamma^{(s)}|}. 
\end{align}
}
Mathematically, the above model is equivalent to
{\small
\begin{align}
    \label{f_formula2}
    f^{(s)}(\cdot) = \langle\Theta(\cdot), \beta^{(s)}\rangle = \sum_{d=1}^{D}\beta^{(s)}_d\theta_d(\cdot), \quad \beta^{(s)}\in \mathbb{R}^{|\Theta|} \text{ and } \lVert \beta^{(s)} \rVert_0 = |\Gamma^{(s)}|. 
\end{align}
}
Formulations (\ref{f_formula1}) and (\ref{f_formula2}) are equivalent because $\beta^{(s)}$ is a sparse coefficient vector which amounts to select $|\Gamma^{(s)}|$ representers from $\Theta$. In (\ref{f_formula1}), both representers and learners are task-specific, where $\Gamma^{(s)}$ involves the representer retrieval and $\alpha^{(s)}$ is a task-specific regression coefficient. In comparison, only sparse regression coefficients $\beta^{(s)}$'s are task-specific in (\ref{f_formula2}). By this means, we cast the representation retrieval framework as a model selection problem, which has been extensively studied \citep{tibshirani1996regression, zou2006adaptive, tibshirani2005sparsity, yuan2006model}. Additionally, more complex learners, such as neural nets, can be adopted to our framework, considering the sparse input features. We refer readers to \citep{wen2016learning, feng2017sparse, lemhadri2021lassonet, fan2024factor} for recent developments in sparse-input neural nets. To keep our introduction concise, we use linear learners in the following discussion. 

To estimate the representers and learners, and pursue the sparsity in $\beta^{(s)}$'s, we can minimize the following loss function for regression problems:
{\small
\begin{align}
    \label{r2_basic_opt}
    \min_{\Theta, \beta^{(1)}, \cdots, \beta^{(S)}} \frac{1}{S}\sum_{s=1}^{S}\frac{1}{n_s}\sum_{i=1}^{n_s}(y^{(s)}_{i} - \langle\Theta(\mathbf{x}_i^{(s)}), \beta^{(s)}\rangle)^2 + \sum_{s=1}^{S}\lambda^{(s)}\lVert \beta^{(s)}\rVert_1,
\end{align}
}
which is termed as representation retrieval ($R^2$) learning. By default, we consider representers $\theta_d$ as neural nets \citep{goodfellow2016deep} in this work due to its adaptation power to different modalities. Note that $\lambda^{(s)}$ controls the level of sparsity, so varying values of $\lambda^{(s)}$'s are adaptive to each data source. The optimization problem in (\ref{r2_basic_opt}) can be solved alternatively between $\Theta$ and $\beta^{(s)}$'s: Given a fixed $\Theta$, minimizing (\ref{r2_basic_opt}) is equivalent to solving Lasso problem for each data source; given fixed $\beta^{(s)}$'s, the dictionary $\Theta$ can be updated through a gradient-based algorithm. Therefore, the computation of our method is more efficient than optimizing over binary matrices as in \citep{gaynanova2019structural, lock2022bidimensional} to recover the partially sharing structure. To handle the classification problem, we can replace the least square loss with cross entropy loss without further modification.

In fact, the proposed model (\ref{f_formula2}) covers many existing methods as special cases. For each single task, model (\ref{f_formula2}) amounts to sparse multiple kernel learning \citep{koltchinskii2010sparsity} if $\theta_d$ corresponds to different kernels. In the simple scenario that $\theta_d(\mathbf{x}) = \mathbf{x}_d$ and $D = p$, our model is equivalent to well-studied sparse linear models for multi-task learning \citep{lounici2009taking, jalali2010dirty, craig2024pretraining}. If $\theta_d$'s are linear mappings, our model is exactly a dictionary learning model for multi-task learning \citep{maurer2013sparse}. Our proposal is also closely related to mixture-of-expert for multi-task learning \citep{ma2018modeling}, where each representer can be treated as an expert.

\subsection{Selective integration penalty}
\label{subsec: sip}

In this section, we introduction the notion of \emph{integrativeness} of representer. Intuitively, data integration happens when the same representer is retrieved in $\beta^{(s)}$ by multiple tasks. Motivated by this, we define the integrativenss of a representer as follows: 
{\small
\begin{align}
    \label{gamma_formula}
    \gamma_{d} = \sum_{s=1}^{S}\mathbb{I}(\beta_d^{(s)} \neq 0),
\end{align}
}
where $0 \le \gamma_d \le S$ counts the number of data sources that retrieves the representer $\theta_d$ for prediction. Theoretically, we will show that more integrative representers lead to the sharper excess risk bound in Section \ref{sec: theory}. However, the optimization problem in (\ref{r2_basic_opt}) does not necessarily favor more integrative representers. Therefore, we introduce additional structural assumptions in the $R^2$ learning for better control of integrativeness to further improve the prediction.

\begin{figure}[!t]
    \centering
    \centering
     \begin{subfigure}[b]{0.3\textwidth}
         \centering
         \raisebox{20mm}{\begin{tikzpicture}
\begin{axis}[
    width=5cm,
    height=5cm,
    xlabel={$\gamma_d$},
    ylabel={$\mathcal{P}$},
    xmin=0, xmax=4,
    ymin=0, ymax=1.5,
    axis lines=middle,
    samples=100,
    domain=0:4,
    ymajorgrids=true,
    xmajorgrids=true,
    grid style=dashed,
    xlabel style={at={(axis description cs:0.5,-0.15)}, anchor=north},
    xtick={1, 2.5, 4},
    xticklabels={$1$, $\cdots$, $S$}
]
\addplot[
    domain=0:4,
    thick,
    color=black,
]
{max(min(4/3-x/3, 1), 0)};

\addplot[
    domain=0:4,
    dashed, thick,
    color=blue,
]
{sqrt(1-x/4)};
\end{axis}
\end{tikzpicture}}
         \caption*{(a)}
     \end{subfigure}
     \hfill
     \begin{subfigure}[b]{0.65\textwidth}
         \centering
         \includegraphics[width=0.55\linewidth]{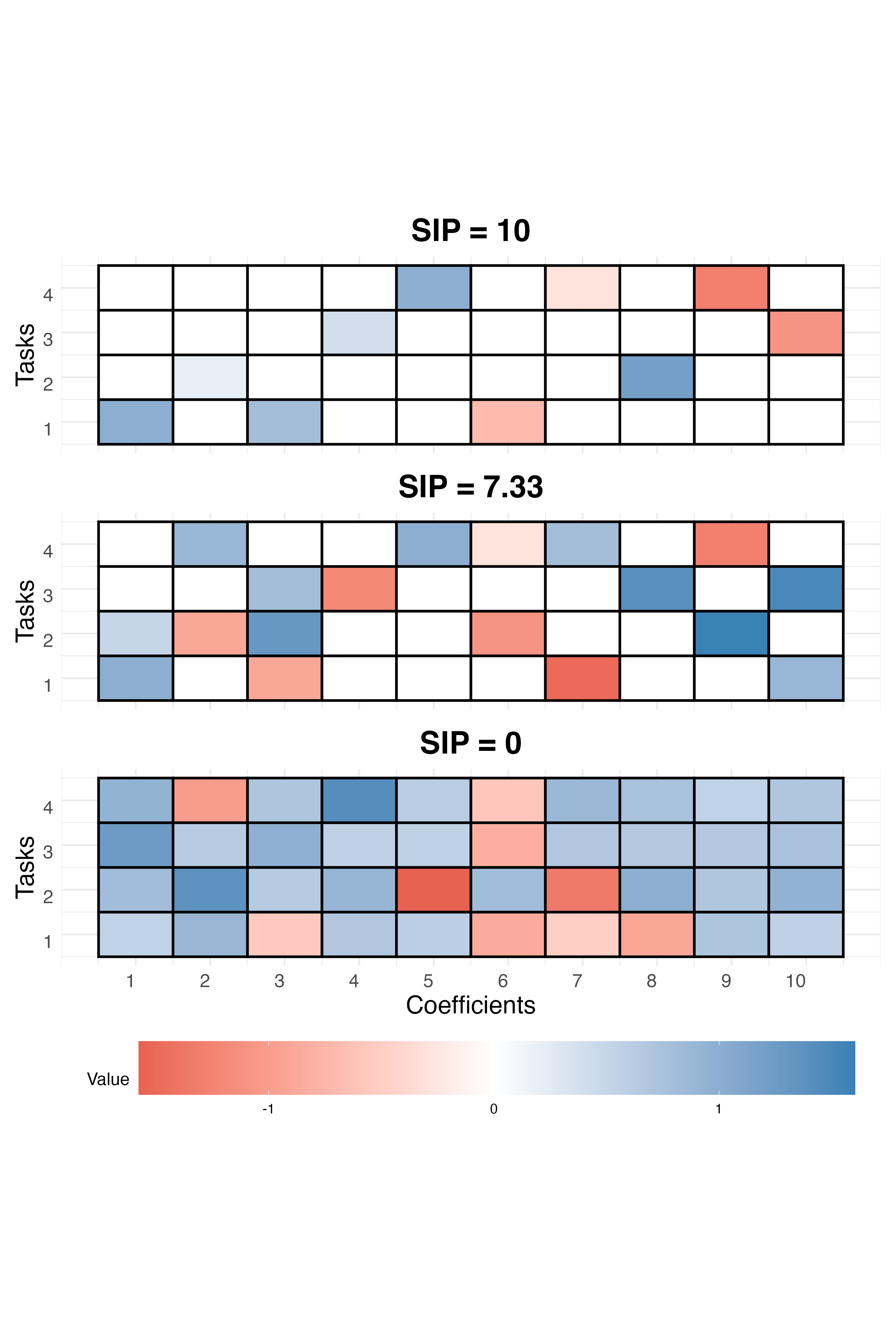}
         \caption*{(b)}
    \end{subfigure}
    \caption{\small (a): Black solid line: curve of SIP formula (\ref{sip_formula}) for a single $\gamma_d$; blue dashed line: curve of $\sqrt{1-\gamma_d/S}$, which will be introduced in Section \ref{sec: theory}. (b): Illustrations of regression coefficients $\beta_d^{(s)}$ under $\mathcal{P} = D$, $0 < \mathcal{P} < D$ and $\mathcal{P} = 0$ scenarios. In this example, we consider $D = 10$ representers, and $S = 4$ tasks.}
    \label{fig: sip_visual}
\end{figure}

Formally, we propose the following \emph{Selective Integration Penalty (SIP)} to encourage the integrativeness of representers:
{\small
\begin{align}
    \label{sip_formula}
    \mathcal{P}(\beta^{(1)}, \cdots, \beta^{(S)}) = \sum_{d=1}^{D}\min(1, \frac{S - \gamma_d}{S-1}),
\end{align}
}
where $0\le \mathcal{P}(\cdot) \le D$. For each single $\gamma_d$, SIP truncates at $\gamma_d = 1$, in that the penalty is identical to $1$ for $0\le \gamma_d \le 1$; since any $0 \le \gamma_d \le 1$ do not integrate any information across tasks, therefore they should be penalized at the same level. When $\mathcal{P}(\cdot) = 0$, it indicates that $\beta_d^{(s)} \neq 0$ for all $d$ and $s$, showing that all the representers are retrieved by all $S$ data sources. On the contrary, if $\mathcal{P}(\cdot) = D$, then $\gamma_d \le 1$ for all $d$, suggesting that each representer is retrieved at most once. In other words, different tasks do not share any common representers. In addition, if $0 < P(\cdot) < D$, a partially sharing structure is encoded in $\beta^{(s)}$. Due to the non-smoothness of the indicator function in (\ref{gamma_formula}), it is intractable to minimize $\mathcal{P}(\cdot)$. Alternatively, we adopt the truncated-$\ell_1$ function \citep{pan2013cluster, wu2016new} as a continuous relaxation of the indicator function in (\ref{gamma_formula}):
{\small
\begin{align}
     \tilde{\gamma}_{d}(\tau) = \sum_{s=1}^{S}\min(1, \frac{|\beta_d^{(s)}|}{\tau}), \notag
\end{align}
}
where $0 \le \tilde{\gamma}_d(\tau) \le S$. Based on the above derivation, the loss function of our $R^2$ learning with SIP is
{\small
\begin{align}
    \label{r2_opt}
    \min_{\Theta, \beta^{(1)}, \cdots, \beta^{(S)}} \frac{1}{S}\sum_{s=1}^{S}\frac{1}{n_s}\sum_{i=1}^{n_s}\lVert y^{(s)}_{i} - \langle\Theta(\mathbf{x}_i^{(s)}), \beta^{(s)}\rangle\rVert_2^2 + \sum_{s=1}^{S}\lambda_1^{(s)}\lVert \beta^{(s)}\rVert_1 + \lambda_2\mathcal{P}_{\tau}(\beta^{(1)}, \cdots, \beta^{(S)}),
\end{align}
}
where $\mathcal{P}_{\tau}(\cdot)$ is the smoothed SIP by replacing $\gamma_d$ with $\tilde{\gamma}_d(\tau)$. Here, two penalties play different roles: the $\ell_1$ penalty in the second term of (\ref{r2_opt}) targets each individual task, in order to select informative representers for each specific task. SIP, on the other hand, acts on regression coefficients across tasks, and seek to improve the overall integrativeness of representers. We employ the Adam \citep{kingma2014adam} optimizer in \texttt{pytorch} to minimize the (\ref{r2_opt}) by alternating between $\Theta$ and $\beta$.

Our proposed $R^2$ learning with SIP offers unique advantages to multi-task learning. Specifically, we allow practitioners to choose flexible representers that are adaptive to their data structure. Since our SIP is defined directly on learners, representers can be different models with varying numbers of parameters. For example, linear mappings can capture linear structures of original data, while neural nets can capture high-order nonlinear structures, and we can combine linear and nonlinear representers in the dictionary as representer candidates for different data sources. This is advantageous for multi-source data where each data type could have varying complexity of structures. In contrast, most existing integrative penalties \citep{tang2016fused, duan2023adaptive, tian2023learning} require model parameters comparable across tasks to apply the penalties, which could be stringent in multi-source data integration. In particular, when multi-source data have different modalities (e.g, some of them have images, while others have text data), it is ineffective to employ the same model for all sources.

\subsection{Blockwise Representation Retrieval Learning}
\label{subsec: BR2learn}

In this section, we propose the Blockwise Representation Retrieval ($BR^2$) learning, which extends the $R^2$ learning to multi-modality data with blockwise missingness.

For ease of our discussion, we first introduce the essential notations for multi-modality data. In total, there are $M$ modalities that are observed by at least one data source. The missing indicators, $\mathbb{I}_m^{(s)}$ for $s=1,\cdots, S$, $m=1\cdots, M$, indicate whether the $s$th data source observing the $m$th modality. That is, $\mathbb{I}_m^{(s)} = 1$ if the $m$th modality is observed in the $s$th data source, and $\mathbb{I}_m^{(s)} = 0$ otherwise. Further, we define $\mathcal{O}_m = \{s: \mathbb{I}_m^{(s)} = 1\}$ as the set of data sources observing the $m$th modality; Similarly, we denote $\mathcal{O}^{(s)} = \{m: \mathbb{I}_m^{(s)} = 1\}$ as the observed modalities of the $s$th data source. For instance, the upper block in Figure \ref{fig: special_cases} presents a $S=5$ and $M=5$ example. As an illustration, $\mathcal{O}^{(1)} = \{1, 2, 4, 5\}$ and $\mathcal{O}_2 = \{1, 3, 4\}$ in this example.

For real applications with blockwise missingness, the missing mechanism is usually unknown and unverifiable. Due to potential covariates shift of marginal distribution of each modality or joint distribution of several modalities, imputation approaches, which learn conditional distribution from other data sources to impute missing modalities for certain data sources, can introduce significant bias. First of all, missing completely at random (MCAR) is unlikely to hold under covariates shift. Second, under missing at random (MAR) or missing not at random (MNAR) mechanisms, the conditional distribution of one modality given other modalities are different across data sources, which invalidate the aforementioned imputation approaches. Therefore, it is of great interest to investigate whether we can improve predictive performance without imputation.

In the following, we propose the \emph{Blockwise Representation retrieval} ($BR^2$) learning, which directly learns the function mappings from observed modalities to responses without imputation. Instead of imputing missing blocks and applying $R^2$ learning to the imputed data, we aim to borrow as much information as we can given the observed data. Specifically, we introduce representer dictionary for each single modalities. To accommodate covariates shift among data sources for each modality, a subset of representers are retrieved as informative ones to predict the response. Later, we concatenate retrieved representers from each dictionary as the latent representation of each data source. Again, we introduce source-specific sparse-induced learners for each modality to handle posterior drift. Formally, we define modality-specific representer dictionaries as $\Theta_{m} = \{\theta_{m; 1}, \cdots, \theta_{m; D_m}\}, m = 1, \cdots, M$, and each representer is a function mapping $\theta_{m; d}: \mathcal{X}_m \rightarrow \mathbb{R}$ as in $R^2$ learning. Given this setup, data sources $s \in \mathcal{O}_{m}$ share the $m$th dictionary, which enables us to borrow information across data sources for each modality. The function mapping $f^{(s)}: \mathcal{X}^{(s)} \rightarrow \mathcal{Y}^{(s)}$ is then modeled as
{\small
\begin{align}
    \label{f_formula3}
    f^{(s)}(\cdot) = \sum_{m=1}^{M}\mathbb{I}_m^{(s)}\langle\Theta_m(\cdot), \beta^{(s)}_m\rangle, \quad \beta^{(s)}_m \in \mathbb{R}^{|\Theta_m|} \text{ and } \lVert\beta_m^{(s)}\rVert_0 = |\Gamma^{(s)}_m|,
\end{align}
}
where $\Gamma_m^{(s)}$ is the subset of retrieved representers from the $m$th modality dictionary $\Theta_m$ for the $s$th data source. In (\ref{f_formula3}), the missing indicator $\mathbb{I}_m^{(s)}$ indicates that if the $m$th modality is unobserved in the $s$th data source, then the $m$th modality has no impact on the response. 

Like linear models, dictionaries $\Theta_{m}$'s represent the main effects of the $m$th modality to responses, while the interaction or association among modalities are ignored. To ease this constraint, there are two options to incorporate the association among modalities in modeling $f^{(s)}$. First, we can introduce interaction dictionaries $\Theta_{m_1, m_2, \cdots, m_w}$, where $m_1, \cdots, m_w \in \{1, \cdots, M\}$ represents the $w-$way interaction among modalities. Similarly, each interaction dictionary $\Theta_{m_1, \cdots, m_w}$ is associated with sparse learner $\beta_{m_1, \cdots, m_w}^{(s)}$ for each source. Second, we may choose more complex learners, such as sparse-input neural nets \citep{feng2017sparse, lemhadri2021lassonet, fan2024factor} to learn the interaction among modalities. Again, we will only consider the modality-specific dictionaries and linear learners in the later theoretical analysis and numerical experiments.

Similar to the $R^2$ learning, the representers and learners in $BR^2$ learning can be estimated by minimizing the following loss function for regression problems:
{\small
\begin{align}
    \label{br2_opt}
    \min_{\Theta, \beta} \frac{1}{S}\sum_{s=1}^{S}\frac{1}{n_s}\sum_{i=1}^{n_s}\lVert y^{(s)}_{i} -\sum_{m=1}^{M}\mathbb{I}_m^{(s)}\langle\Theta_m(\mathbf{x}_{i; m}^{(s)}), \beta_m^{(s)}\rangle\rVert_2^2 &+ \sum_{s=1}^{S}\sum_{m=1}^{M}\lambda_{m; 1}^{(s)}\lVert \beta_m^{(s)}\rVert_1 \notag,\\
    &+\sum_{m=1}^{M}\lambda_{m;2}\mathcal{P}(\beta_{\mathcal{O}_m}) 
\end{align}
}
and squared loss can be replaced with other loss functions, such as cross entropy loss for classification problems. Since $\Theta_m$ is modality-specific dictionary and can only be trained with data sources observing the $m$the modality, its integrativenss is defined as $\gamma_{m;d} = \sum_{s\in \mathcal{O}_m}\mathbb{I}(\beta^{(s)}_{m;d} \neq 0)$. As a result, the SIP for the $m$th modality is only effective for data sources in $\mathcal{O}_m$, $\beta_{\mathcal{O}_m}$, which improves the modality-specific integrativeness of representers in $\Theta_m$..

Compared with supervised learning for individual sources ignoring the missing modalities, our $BR^2$ learning improves the integrativeness of each modality-specific representations that borrow information from other sources observing the same modalities, which performs at least comparably well to learning with individual sources. In comparison, supervised learning based on imputed values could diminish the predictive power if imputation models are mis-specified, which may perform even worse than learning with individual sources. Empirical comparisons among these methods are provided in Section \ref{subsec: BR2_sim}.

\section{Theoretical Results}
\label{sec: theory}

In this section, we study the generalization performance of the $R^2$ framework. In particular, we first establish the excess risk bound of $R^2$ learning and then extend to $BR^2$ learning. The theoretical results in this section will reveal that the generalization performance of $R^2$ learning is impacted by the \emph{representer retrieval pattern}. In words, given sparse constraints on representer retrieval, it is preferable to have \emph{less representers shared by more data sources}, than \emph{more representers shared by less data sources}. This finding resonates with the proposed SIP in Section \ref{subsec: sip}. In the end, the empirical advantage of SIP over the theoretical upper bound will be discussed.

First, we formalize the function classes of (\ref{f_formula2}) as follows: Suppose each representer $\theta_d \in \mathcal{G}_d$, where $\mathcal{G}_d$ is certain function class for the $d$th representer, such as linear function or neural nets with norm constraints. As for sparse learners, we consider the sparse regression coefficient class: $\beta^{(s)} \in \mathcal{H}^{(s)} = \{\beta: \lVert \beta\rVert_1 \le \alpha^{(s)}\}$. For brevity, we denote $\mathcal{G}$ as the function space for $\Theta = (\theta_1, \cdots, \theta_D)$ and $\mathcal{H}$ as the space for $\beta = (\beta^{(1)}, \cdots, \beta^{(S)})$. Then, the function classes of each $\mathcal{F}^{(s)}$ is characterized by $\mathcal{G}_1, \cdots, \mathcal{G}_D$ and $\mathcal{H}^{(s)}$: 
\begin{equation}
    \mathcal{F}^{(s)} = \{f^{(s)}: f^{(s)}(\cdot) = \sum_{d=1}^{D}\beta_d^{(s)}\theta_d(\cdot), \beta^{(s)} = (\beta^{(s)}_1, \cdots, \beta^{(s)}_D) \in \mathcal{H}^{(s)}, \theta_d\in \mathcal{F}_d\}. \notag
\end{equation}
The function classes $\mathcal{G}_d$'s and $\mathcal{H}^{(s)}$'s admits the partially sharing structures across data sources in that each data source retrieves some of representers from the dictionary, which relaxes the fully sharing structure studied in \citep{tripuraneni2021provable, watkins2023optimistic}.

Assuming each data source is endowed with distribution $\mu^{(s)}$, and we can observe $n_s$ i.i.d samples $(\mathbf{x}_i^{(s)}, y_i^{(s)})$, $i=1, \cdots, n_s$ from $\mu^{(s)}$. Then, the empirical risk minimization of $R^2$ learning can be formulated as follows:
\begin{equation}
    \min_{\Theta\in\mathcal{G}, \beta\in \mathcal{H}}\hat{\mathcal{R}}(\Theta, \beta) := \min_{\Theta\in\mathcal{G}, \beta\in \mathcal{H}}\frac{1}{S}\sum_{s=1}^{S}\frac{1}{n_s}\sum_{i=1}^{n_s}\ell\bigg\{y_i^{(s)}, \sum_{d=1}^{D}\theta_d(\mathbf{x}_i^{(s)})\beta_d^{(s)}\bigg\}, \notag
\end{equation}
where the empirical risk minimizer is denoted as $\hat{\Theta}$ and $\hat{\beta}$. For simplicity, we assume $n = n_1 = \cdots = n_S$ in the following theoretical development. Analogously, the population risk is defined as 
\begin{equation}
    \mathcal{R}(\Theta, \beta) = \frac{1}{S}\sum_{s=1}^{S}\mathbb{E}_{(\mathbf{x}^{(s)}, y^{(s)})\sim \mu^{(s)}}\bigg[\ell\bigg\{y^{(s)}, \sum_{d=1}^{D}\theta_d(\mathbf{x}^{(s)})\beta_d^{(s)}\bigg\}\bigg], \notag
\end{equation}
and $\Theta^*, \beta^*$ denote the minimizers of population risk. Further, we denote the population risk given true $f^{(s)}$ as $\mathcal{R}(f) = \frac{1}{S}\sum_{s=1}^{S}\mathbb{E}_{(\mathbf{x}^{(s)}, y^{(s)})\sim \mu^{(s)}}\bigg[\ell\bigg\{y^{(s)}, f^{(s)}(\mathbf{x}^{(s)})\bigg\}\bigg]$. Since the postulated function space $\mathcal{F}^{(s)}$ may not contain the true function $f^{(s)}$, $\mathcal{R}(\theta^*, \beta^*) - \mathcal{R}(f)$ signifies the approximation error of $\mathcal{F}^{(s)}$ towards $f^{(s)}$. Our target is the excess risk between the empirical risk minimizer and the true functions $f^{(s)}$'s, defined as
\begin{align}
    \mathcal{E}(\hat{\Theta}, \hat{\beta}) &= \mathcal{R}(\hat{\Theta}, \hat{\beta}) - \mathcal{R}(f) \notag \\
    &= \underbrace{\mathcal{R}(\hat{\Theta}, \hat{\beta}) - \mathcal{R}(\Theta^*, \beta^*)}_{\text{estimation error}} + \underbrace{\mathcal{R}(\Theta^*, \beta^*) - \mathcal{R}(f)}_{\text{approximation error}}. \notag
\end{align}
In this decomposition, the approximation error depends on several factors, such as the choice of $\mathcal{G}_d$'s, and properties of true function $f^{(s)}$'s and so on. In the following analysis, we focus on the estimation error from finite sample, given any choice of $\mathcal{G}_d$ with some mild regularity conditions. Now, we proceed with some essential assumptions:
\begin{assumption} 
    \label{assumption1}
    \begin{enumerate}[label=(\alph*)]
        \item (Lipschitz loss) The loss function: $\ell(\cdot, \cdot)$ is bounded, and $\ell(y, \cdot)$ is $L$-Lipschitz for all $y \in \mathcal{Y}$.
        \item (Boundedness) The representer is bounded for $d \in [D]$: $\sup_{\mathbf{x} \in \mathcal{X}}|\theta_d(\mathbf{x})| \le B$ for any $\theta_d \in \mathcal{F}_d$. In addition, if $\mathcal{Y}=\mathbb{R}$, we assume $|y| \le B$.
    \end{enumerate}
\end{assumption}
\begin{remark}
     Assumptions (a) and (b) are standard assumptions in learning theory to control the complexity of function classes, which can be satisfied with bounded $\mathcal{Y}$ and $\mathcal{G}$ under common adopted loss functions, such as least square loss and cross entropy loss. 
\end{remark}

In the following, we show that the complexity of the representer dictionary is directly impacted by $\gamma_d$, the integrativeness of the $d$th representer.

\begin{lemma}
\label{lemma1}
For any $\beta\in \mathcal{H}$ and fixed realizations $(\mathbf{x}_i^{(s)}, y_i^{(s)})_{i=1}^{n}$, we define 
{\small
\begin{align}
    \mathbf{G}_{\beta}(\boldsymbol{\sigma}) = \sup_{\Theta\in\mathcal{G}}\frac{1}{nS}\sum_{s=1}^{S}\sum_{i=1}^{n}\sigma_{s, i}\bigg[\sum_{d=1}^{D}\theta_d(\mathbf{x}_i^{(s)})\beta_d^{(s)}\bigg], \notag
\end{align}
}
where $\sigma_{s, i}$ are i.i.d Rademacher random variables, then we have:
{\small
\begin{align}
\label{lemma1_bound1}
\mathbb{E}_{\boldsymbol{\sigma}}\mathbf{G}_{\beta}(\boldsymbol{\sigma}) &\le \alpha\sqrt{\frac{C_{\Theta}}{nS^2}}\sum_{d=1}^{D}\sqrt{\gamma_d},
\end{align}
}
with $\gamma_d = \sum_{s=1}^{S}\mathbb{I}(\beta_d^{(s)} \neq 0)$. The constant $C_{\Theta}$ is the uniform upper bound $C_{\Theta} \ge C_{\theta_d}$ for all $d$, where $C_{\theta_d}$ is the Rademacher complexity of function class $\mathcal{G}_d$. And $\alpha \ge \alpha^{(s)}$ is the uniform upper bound for all $\alpha^{(s)}$.
\end{lemma}

The quantity $\mathbb{E}_{\boldsymbol{\sigma}}\mathbf{G}_{\beta}(\boldsymbol{\sigma})$, analogous to the Rademacher complexity, quantifies the capacity of the representation dictionary to fit random noise given any fixed $\beta \in \mathcal{H}$. A similar measure has been studied for linear dictionaries in \citep{maurer2013sparse}, in which they obtained a looser upper bound that relies on all $D$ representers. In contrast, we refine the analysis by leveraging the support of the given $\beta$ in Lemma \ref{lemma1}, which leads to the $\sum_{d=1}^{D}\sqrt{\gamma_d}$ term in (\ref{lemma1_bound1}). We refer readers to Appendix \ref{app_sec: proof} for more detailed technical derivation. 

We first instantiate the upper bound (\ref{lemma1_bound1}) with two extreme cases for interpretation. If $\alpha_s$'s are large enough for all data sources, or $\gamma_d = S$ for all representers, the upper bound is $\alpha\sqrt{\frac{C_{\Theta}D^2}{nS}}$, which utilizes all data sources to learn all representers, and matches the order of $O(\sqrt{1/nS})$ in existing multi-task learning literature \citep{tripuraneni2021provable}. In this regime, large $D$ incurs potential overfitting that worsen the estimation error, signifying the drawback of the fully sharing structure. On the other hand, if $\gamma_d=1$ for all representers, the dictionary size $D$ is at least $S$, which leads the upper bound to be at least $\alpha\sqrt{\frac{C_{\Theta}}{n}}$, indicating no benefit of integrating all data sources to learn $\Theta$. In general, the complexity of $\Theta$ lies between the above two extreme cases in that shared representers can lower the complexity. Due to the concavity of $\sum_{d=1}^{D}\sqrt{\gamma_d}$, the upper bound in (\ref{lemma1_bound1}) is lower when non-zero $\gamma_d$'s are concentrated in few representers given fixed function class $\mathcal{G}$. In other words, \emph{less representers shared by more data sources} lowers the upper bound in (\ref{lemma1_bound1}) than \emph{more representers shared by less data sources}. In the following, we obtain the excess risk bound for $R^2$ learning, and provide more detailed discussion on the role of integrativeness.

\begin{theorem}
    \label{theorem1}
    Suppose Assumption \ref{assumption1} holds, and the dictionary size is fixed as $D$. Constants $C_{\Theta}$ and $\alpha$ are defined as in Lemma \ref{lemma1}. Then for empirical risk minimizer $\hat{\Theta}$ and $\hat{\beta}$, with probability $1-\delta$, we have
    {\small
    \begin{align}
        \label{theorem1_bound}
        \mathcal{R}(\hat{\Theta}, \hat{\beta}) - \mathcal{R}(\Theta^*, \beta^*) \le 4L\alpha\sqrt{\frac{C_{\Theta}}{nS^2}}\sup_{\beta\in \mathcal{H}}\sum_{d=1}^{D}\sqrt{\gamma_d} + 4\alpha B L\sqrt{\frac{2\log(2D)}{n}} + B\sqrt{\frac{2\log(2/\delta)}{nS}}.
    \end{align}
    }
\end{theorem}

In this upper bound, the first term in (\ref{theorem1_bound}) is from Lemma \ref{lemma1}, quantifying the cost of learning the representation dictionary. The second term of (\ref{theorem1_bound}) arises from learning $\beta^{(s)}$ for each task, which scales with $O(n^{-1/2})$. And the last term of (\ref{theorem1_bound}) signifies the cost of learning from random samples and converges fast as either $n$ or $S$ increases. 

For the first term, the cost of learning representation dictionary, depends on the term $\sup_{\beta\in \mathcal{H}}\sum_{d=1}^{D}\sqrt{\gamma_d}$, which motivates us to impose additional constraint on $\mathcal{H}$ (denoted as $\mathcal{H}'$ temporarily) to lower $\sup_{\beta\in \mathcal{H}'}\sum_{d=1}^{D}\sqrt{\gamma_d}$. Note that minimizing $\sum_{d=1}^{D}\sqrt{\gamma_d}$ is indeed equivalent to minimizing $\sum_{d=1}^{D}\sqrt{1-\gamma_d/S}$, where the curve $\sqrt{1-\gamma_d/S}$ is shown in blue dashed line in Figure \ref{fig: sip_visual}. Further, SIP (black solid line in Figure \ref{fig: sip_visual}) can be viewed as an piece-wise linear approximation of $\sqrt{1-\gamma_d/S}$. 

Empirically, we prefer to minimize SIP instead of $\sum_{d=1}^{D}\sqrt{\gamma_d}$ or $\sum_{d=1}^{D}\sqrt{\tilde{\gamma}_d}$ due to the following reasons: the gradient of $\sqrt{1-\gamma_d/S}$ is approaching negative infinity when $\gamma_d \to S$, it not only strongly encourages fully sharing structure, but also leads to unstable training when $\gamma_d$ approaching $S$. On the contrary, the gradient of SIP with respect to $\gamma_d$ is constant for $\gamma_d > 1$, fairly encourage the representer to be shared by more representers. Combined with $\ell_1$ penalty for each data source, and data fitting loss functions, partially sharing structure can be easily captured. Further, the constant gradient leads gradient-descent type algorithm more stable in optimization.

Mathematically, minimizing (\ref{r2_opt}) is equivalent to impose constraint $\sum_{d=1}^{D}\sqrt{\tilde{\gamma}_d(\tau)} \le \eta(\tau)$ on $\mathcal{H}$, for some $\eta(\cdot)$, which will be denoted as $\mathcal{H}_{\eta, \tau} = \{\sum_{d=1}^{D}\sqrt{\tilde{\gamma}_d(\tau)}\le \eta(\tau), \beta \in \mathcal{H}\}$ hereafter. In the following Corollary, we present the estimation error bound of minimizer of the following empirical risk:
\begin{align}
    \check{\Theta}, \check{\beta} = \argmin_{\Theta\in\mathcal{G}, \beta\in \mathcal{H}_{\eta, \tau}}\hat{\mathcal{R}}(\Theta, \beta) := \argmin_{\Theta\in\mathcal{G}, \beta\in \mathcal{H}_{\eta, \tau}}\frac{1}{S}\sum_{s=1}^{S}\frac{1}{n_s}\sum_{i=1}^{n_s}\ell\bigg\{y_i^{(s)}, \sum_{d=1}^{D}\theta_d(\mathbf{x}_i^{(s)})\beta_d^{(s)}\bigg\}. \notag
\end{align}

\begin{corollary}
    \label{corollary1}
    Suppose Assumption \ref{assumption1} holds, and the dictionary size is fixed as $D$. Constants $C_{\Theta}$ and $\alpha$ are defined as in Lemma \ref{lemma1}. Then for empirical risk minimizer $\check{\Theta}$ and $\check{\beta}$, with probability $1-\delta$, we have
    {\small
    \begin{align}
        \label{corollary1_bound}
        \mathcal{R}(\check{\Theta}, \check{\beta}) - \mathcal{R}(\Theta^*, \beta^*) \le 4L\alpha\sqrt{\frac{C_{\Theta}}{nS^2}}\eta(\tau) + 4L\tau\sqrt{\frac{C_{\Theta}D^2}{nS}} + 4\alpha B L\sqrt{\frac{2\log(2D)}{n}} + B\sqrt{\frac{2\log(2/\delta)}{nS}}.
    \end{align}
    }
\end{corollary}

In the continuous relaxation of SIP with truncated $\ell_1$ function, the parameter $\tau$ serves as a tuning parameter which specifies the threshold at which a coefficient $\beta_d^{(s)}$ is regarded as meaningfully non-zero. Therefore, the first term indicates the complexity of learning the representer when regression coefficients are meaningfully non-zero ($|\beta_d^{(s)}| > \tau$), and the second term indicates the complexity of learning the representer when $|\beta_d^{(s)}| \le \tau$. When $\tau\to 0$, the second term vanishes, which indicates that the first term captures the main complexity of learning $\Theta$ and the second terms captures the additional cost of approximating $\gamma_d$ with $\tilde{\gamma}_d(\tau)$. In addition, since $\tau$ is usually far less than $\alpha$, the first term dominates the second term in the above upper bound.

There are several trade-offs in the excess risk bound. First of all, the first term in (\ref{corollary1_bound}) is monotonically decreasing with respect to $\tau$, while the second term is monotonically increasing with respect to $\tau$. Consequently, choosing a proper $\tau$ can effectively balance these two terms. Second, the dictionary size $D$, sparsity constraint $\alpha$ and representer complexity $C_{\Theta}$ determine the model capacity jointly. In particular, the complexity of representer $C_{\Theta}$ impact the approximation error and the complexity of learning $\Theta$ (\ref{corollary1}). In general, a simple representer model (small $C_{\Theta}$) would potentially worsen the approximation error and improve the generalization error, and vice versa. The sparsity constraint $\alpha$ is involved in the approximation error, and the complexity of learning both representer and regressors in the estimation error. A sparse regressor (small $\alpha$) would potentially worsen the approximation error and improve the estimation error, and vice versa. The dictionary size $D$ is slightly different, the complexity of learning $\Theta$ in (\ref{corollary1_bound}) is robust with respect to $D$. In particular, since SIP encourages to minimize $\sum_{d=1}^{D}\sqrt{\tilde{\gamma}_d(\tau)}$, which leads to many representers with $\tilde{\gamma}_d(\tau) = 0$, they do not increase the complexity of learning $\Theta$. Even though a large $D$ could increase the cost of learning regressors in $O(\sqrt{\log(D)})$, its impact is relatively minimal due to the logarithm. Therefore, in our implementation, we fix the dictionary size $D = 30$ and tune other tuning parameters to optimize the generalization performance.

The theoretical property of $BR^2$ learning can be derived with the similar logic and derivation, which is provided in Appendix \ref{app_sec: br2_theory}.

\section{Simulation Studies}
\label{sec: simulation}

We evaluate the performance of the proposed $R^2$ and $BR^2$ learning approaches under various simulation settings and compare them with several competitive methods. Section \ref{subsec: R2_sim} investigates a multi-task learning (MTL) scenario with fully aligned covariates across data sources, comparing $R^2$ learning against standard MTL methods. Section \ref{subsec: BR2_sim} examines a multi-modality blockwise missing setting, where $BR^2$ learning is compared with methods designed for block-missing data. In both cases, our methods demonstrate superior performance, highlighting their effectiveness in integrating heterogeneous data.

\subsection{Simulations for Multi-task Learning}
\label{subsec: R2_sim}

We compare our method with several baseline and state-of-the-art MTL approaches. The baseline models include Single Task Learning (STL), which fits separate models for each data source using either a linear model (STL-linear) or a feed-forward neural network (STL-NN). Another baseline is Data Pooling (Pooling), which fits a single model across all sources, thereby ignoring any types of heterogeneity. We also use either linear model (Pooling-Linear) or neural net (Pooling-NN) for prediction. These baselines represent two extremes: Pooling is optimal when all tasks are sampled from the same distribution, whereas STL is preferable when no or very few information is shared across data sources.

We also compare our method against prominent multi-task methods from the literature. First, we consider Multi-task Representation Learning (MTRL; \citealt{du2020few, tripuraneni2021provable}), employing a uniform linear (MTRL-Linear) or neural net (MTRL-NN) representer along with source-specific learners. Second, we evaluate the penalized ERM algorithm (pERM; \citealt{tian2023learning}), which allows for similar, though not identical, linear representations across sources and accommodates a subset of outlier sources. Third, Adaptive and Robust Multi-task Learning (ARMUL; \citealt{duan2023adaptive}) is included, as it encourages regression coefficients to conform to a low-rank representation. In all above methods, we specify one hidden layer with 32 neurons as neural net model or representer. The implementation details are provided in Appendix \ref{app_sec: implementation}.

We generate simulation data from $S=20$ data sources as follows:
\begin{align}
{\small
    y^{(s)}_i = \Theta(\mathbf{x}^{(s)}_i)\beta^{(s)} + \epsilon^{(s)}_i, \quad i=1,\cdots, n_s, \quad s=1,\cdots,S,
}
\end{align}
where covariates $\mathbf{x}^{(s)}_i$ follow a multivariate standard normal distribution with dimension $p=30$ and $\epsilon_i^{(s)}$'s are sampled from standard normal distribution. The representer dictionary $\Theta(\cdot) = (\theta_1, \cdots, \theta_{30})$ contains $D=30$ representers, where the first ten representers are linear, the last twenty representers are nonlinear. The detailed specification of these representers can be found in Appendix \ref{app_sec: spec_R2_sim}. Coefficients $\beta^{(s)}$'s are $D=30$-dimensional sparse vector with 5 non-zero entries, where the indices of non-zero are sampled from 30 dimension with the following five different distributions:
\begin{multicols}{2}
\begin{enumerate}
    \item[1.] $(\underbrace{1/5, \cdots, 1/5}_{5\text{ entries}}, \underbrace{0, \cdots, 0}_{25\text{ entries}})$ 
    \item[3.] $(\underbrace{0, \cdots, 0}_{10\text{ entries}}, \underbrace{1/5, \cdots, 1/5}_{5\text{ entries}}, \underbrace{0, \cdots, 0}_{15\text{ entries}})$
    \item[5.] $(\underbrace{1/30, \cdots, 1/30}_{30\text{ entries}})$.
    \item[2.] $(\underbrace{1/10, \cdots, 1/10}_{10\text{ entries}}, \underbrace{0, \cdots, 0}_{20\text{ entries}})$ 
    \item[4.] $(\underbrace{0, \cdots, 0}_{10\text{ entries}}, \underbrace{1/10, \cdots, 1/10}_{10\text{ entries}}, \underbrace{0, \cdots, 0}_{10\text{ entries}})$
\end{enumerate}
\end{multicols}
In particular, the first and second distributions (linear setting) retrieve linear representers, and the third and fourth distributions (nonlinear setting) retrieve nonlinear representers. Further, the first and third distributions specify fully-sharing structure that each data source retrieves the same five representers. On the contrary, the second and fourth distributions allow partially sharing structure that each data source can flexibly retrieve five out of ten representers. The last distribution (mixed setting) allows all data sources to flexibly retrieve any represeneters with uniform probability. On average, each representer will be retrieved only 10/3 times, where information across data sources are hardly borrowed. We also consider two settings for the values of non-zero entries in $\beta^{(s)}$'s: each non-zero entry is sampled from $\mathcal{N}(1, \sigma^2)$ and $\sigma=0, 1$ to reflect the homogeneous and heterogeneous posterior distribution of response given latent representations. In total, we consider $5\times 2=10$ simulation settings, and each simulation setting is repeated 100 times by default.

In the simulation experiment, we set sample size $n_s = 100$ for each source in the training, and $n_s=1000$ for validation, and testing datasets. Model performance is evaluated using the average root mean squared error (RMSE) across all data sources. The predictive performance under nonlinear and mixed settings are presented in Figures \ref{fig: sim1_nonlinear} and \ref{fig: sim1_mixed}, respectively, where linear setting is presented in Appendix \ref{app_sec: add_r2_sim}. Overall, $R^2$ learning with SIP achieves superior or comparable performance under varying settings, especially under partially sharing and posterior drift scenarios. And the effect of SIP for improving predictive performance is especially significant when neural nets are employed as representers.

\begin{figure}[H]
    \centering
    \includegraphics[width=1.0\linewidth]{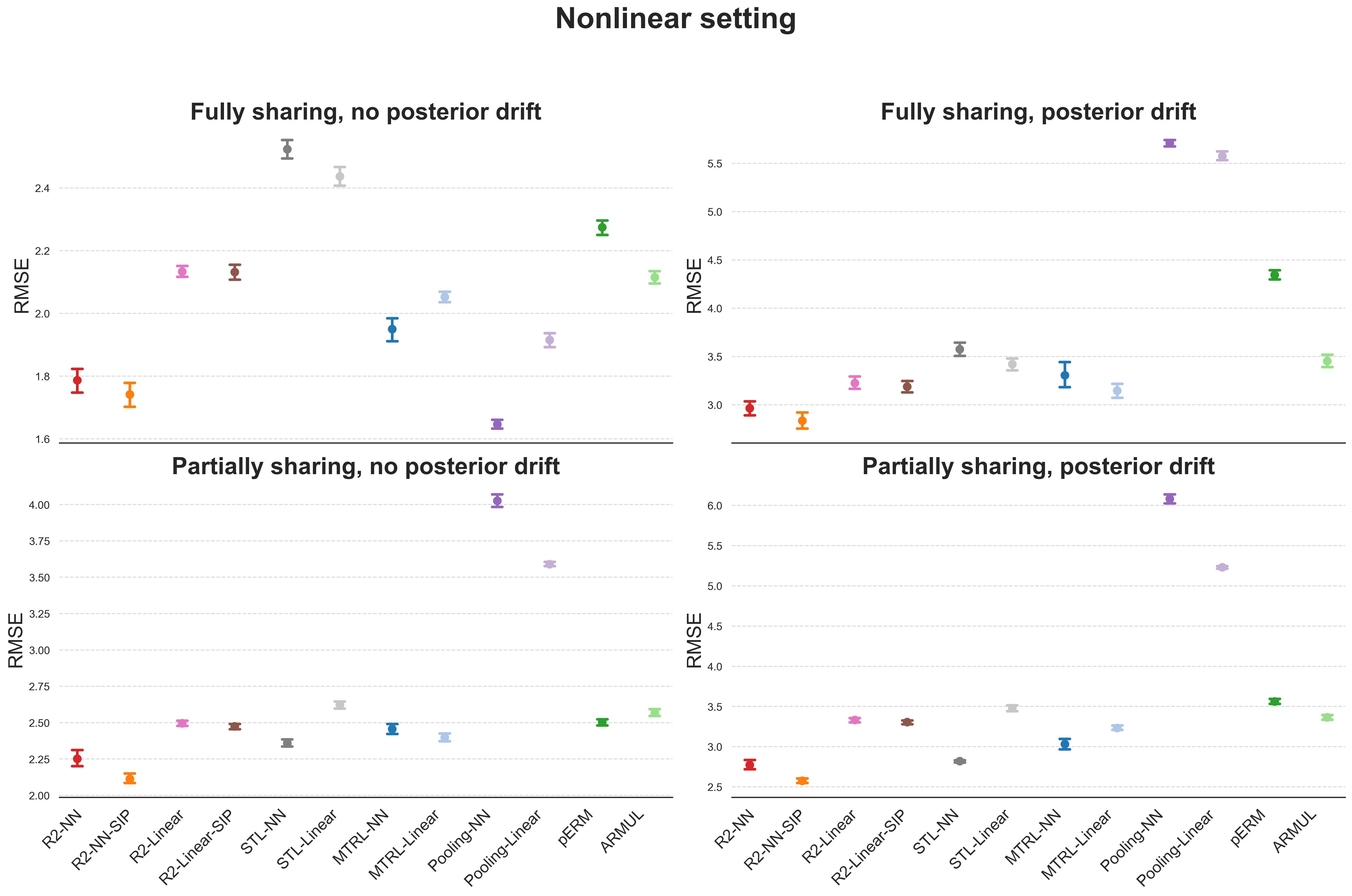}
    \caption{Simulation results for multi-task learning problem under nonlinear setting.}
    \label{fig: sim1_nonlinear}
\end{figure}

Under the nonlinear setting, Pooling-NN achieves the best performance because it can correctly borrow information across all data sources in the homogeneous (fully sharing, no posterior drift) scenario. Beyond that, the $R^2$ learning with neural net representers and SIP outperforms all other competing methods if there is either partially sharing structure or posterior drift across data sources. In the mixed setting, heterogeneity across all data sources is rather significant, in that information across all data sources are hardly integrated unless true data generating procedure is known. As a result, STL-NN is a competitive baseline. Still, $R^2$ learning with neural net representers with SIP can achieve comparable and even better predictive performance.

\begin{figure}[H]
    \centering
    \includegraphics[width=1.0\linewidth]{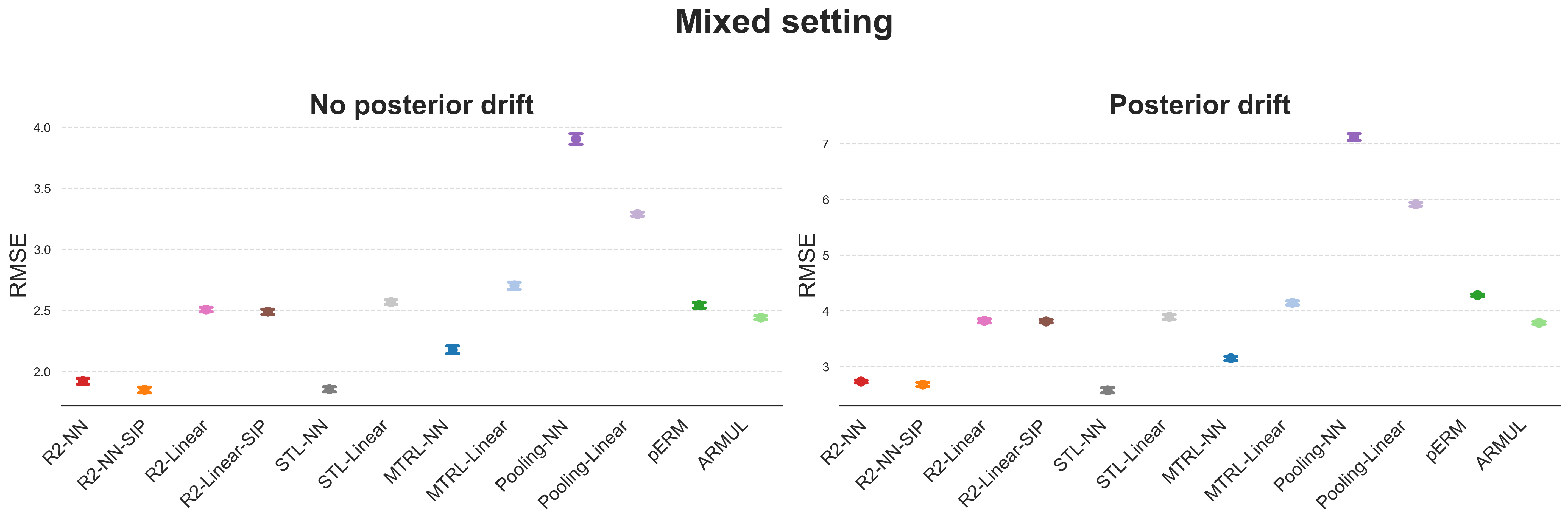}
    \caption{Simulation results for multi-task learning problem under mixed setting.}
    \label{fig: sim1_mixed}
\end{figure}

We provide additional simulation studies in Appendix \ref{app_sec: add_sim} to illustrate the performance of our method in classification problems and varying sample size settings.

\subsection{Simulations for Multi-modality Block-mising Data}
\label{subsec: BR2_sim}

This subsection examines the performance of our proposed $BR^2$ learning method in a multi-modality block-missing data setting. To our knowledge, no existing methods simultaneously address flexible block-missing data, complex covariates shifts, and posterior drift. We compare $BR^2$ learning with a benchmark Single Task Learning (STL) method that uses only the observed modalities, and with DISCOM \citep{yu2020optimal}, which directly predicts without imputing missing modalities by leveraging all available data for estimation of covariance and cross-covariance matrices. Note that the Pooling method is inapplicable here due to missing modalities. Hyperparameters in \citep{yu2020optimal} are selected via grid search in their program, and we still employ \href{https://optuna.org}{\texttt{optuna}} package to tune the hyperparameters in our proposed $BR^2$ learning.

\begin{figure}[H]
    \centering
    \scalebox{1.0}{\begin{tikzpicture}

\node (ph1) at (-1.5, 1.5) {};
\node (ph2) at (-1.5, -2) {};
\node (ph3) at (8, 1.5) {};

\draw[->, thick] (ph1) -- (ph2) node[midway, above, sloped, rotate=180] {\small Data sources};
\draw[->, thick] (ph1) -- (ph3) node[midway, above] {\small Modalities};

\matrix[matrix of nodes, ampersand replacement=\&, nodes={minimum width=5mm, minimum height=5mm, anchor=center, draw}] (L4) at (0, 0) {
|[fill=lightblue3]| \& |[fill=lightpurple3!0]| \&|[fill=lightred4!0]| \& |[fill=orange3!0]| \\
|[fill=lightblue3!0]| \& |[fill=lightpurple3]| \&|[fill=lightred4!0]| \& |[fill=orange3!0]| \\
|[fill=lightblue3!0]| \& |[fill=lightpurple3!0]| \&|[fill=lightred4]| \& |[fill=orange3!0]|\\
|[fill=lightblue3!0]| \& |[fill=lightpurple3!0]| \&|[fill=lightred4!0]| \& |[fill=orange3]|\\
};

\node (L4_caption) at (0, -2) {$L = 1$};

\matrix[matrix of nodes, ampersand replacement=\&, nodes={minimum width=5mm, minimum height=5mm, anchor=center, draw}] (L2) at (3, 0) {
            |[fill=lightblue3]| \& |[fill=lightpurple3]| \&|[fill=lightred4!0]| \& |[fill=orange3!0]| \\
            |[fill=lightblue3!0]| \& |[fill=lightpurple3]| \&|[fill=lightred4]| \& |[fill=orange3!0]|\\
            |[fill=lightblue3!0]| \& |[fill=lightpurple3!0]| \&|[fill=lightred4]| \& |[fill=orange3]| \\
            |[fill=lightblue3]| \& |[fill=lightpurple3!0]| \&|[fill=lightred4!0]| \& |[fill=orange3]| \\
          };
\node (L2_caption) at (3, -2) {$L = 2$};

\matrix[matrix of nodes, ampersand replacement=\&, nodes={minimum width=5mm, minimum height=5mm, anchor=center, draw}] (L3) at (6, 0) {
            |[fill=lightblue3]| \& |[fill=lightpurple3]| \&|[fill=lightred4]| \& |[fill=orange3!0]| \\
            |[fill=lightblue3!0]| \& |[fill=lightpurple3]| \&|[fill=lightred4]| \& |[fill=orange3]| \\
            |[fill=lightblue3]| \& |[fill=lightpurple3!0]| \&|[fill=lightred4]| \& |[fill=orange3]| \\
            |[fill=lightblue3]| \& |[fill=lightpurple3]| \&|[fill=lightred4!0]| \& |[fill=orange3]|\\
          };

\node (L3_caption) at (6, -2) {$L = 3$};
    
\end{tikzpicture}}
    \caption{\label{fig: BR2_sim_missing_pattern} \small Blockwise missing patterns considered in simulation study. $L$ indicates the number of observed modalities for each data source.}
\end{figure}

Figure \ref{fig: BR2_sim_missing_pattern} illustrates the blockwise missing patterns considered in the simulation where \(L\) denotes the number of modalities observed per source. In this experiment, we consider \(S=4\) data sources, \(M=4\) modalities, and set the sample size \(n_s = 200\) for training, validation, and testing. Data are generated according to:
{\small
\begin{align}
y^{(s)} = \sum_{m=1}^{M}\Theta_m(\mathbf{X}_m^{(s)})\beta_m^{(s)} + \epsilon^{(s)}, \notag
\end{align}
}
where modality-specific covariates \(\mathbf{X}_m^{(s)}\) are sampled from a \(q=40\)-dimensional multivariate normal distribution with modality-specific covariance structures (details are provided in Appendix \ref{app_sec: spec_BR2_sim}). For fair comparison, we restrict the representers to the linear case (orthogonal bases in \(\mathbb{R}^{q}\)) as DISCOM is designed for linear models. For each modality, only two representers are active, i.e., \(\lVert \beta_m^{(s)}\rVert_0 = 2\), and the noise term $\epsilon^{(s)}$ are i.i.d samples of standard normal distribution..

\begin{figure}[t]
    \centering
    \includegraphics[width=1.0\linewidth]{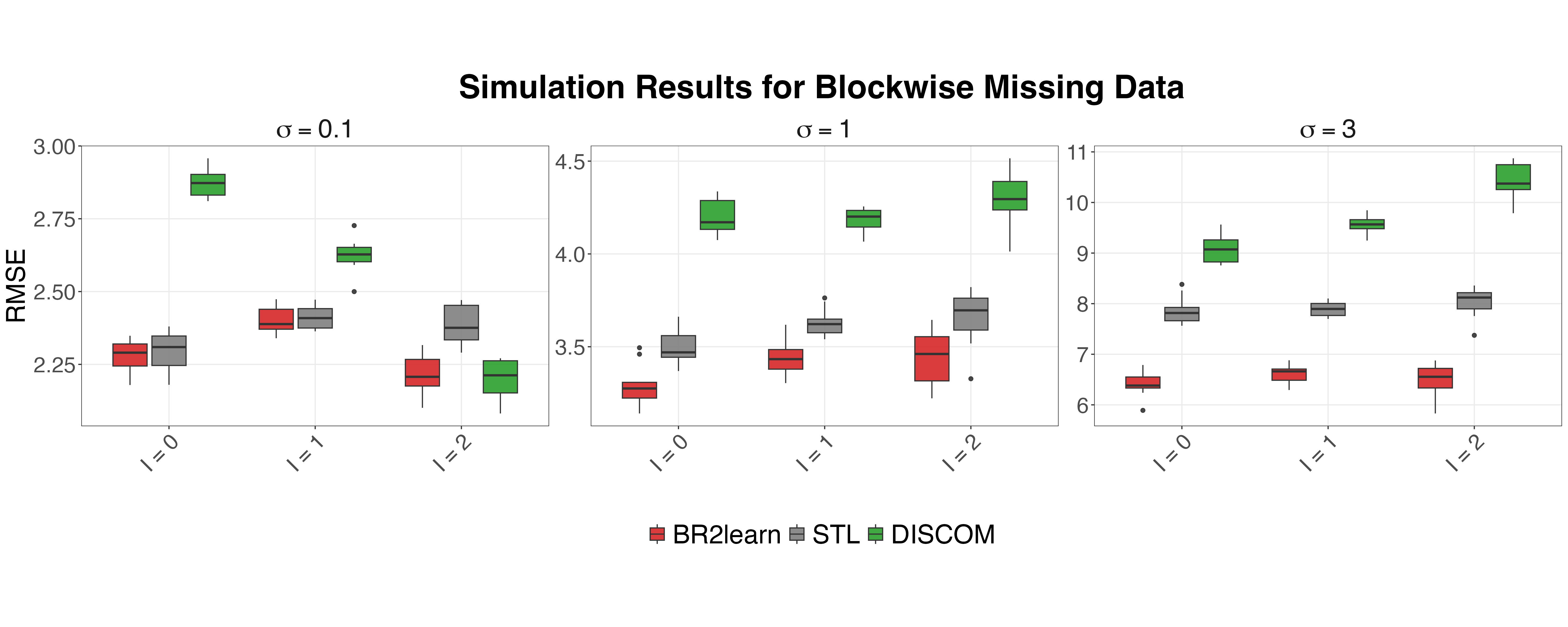}
    \caption{\small Simulation results for blockwise missing learning under varying settings: $I \in \{0, 1, 2\}$, $\sigma \in \{0.1, 1, 3\}$ and $L=3$.}
    \label{fig: sim2_results}
\end{figure}

Further, we consider the number of shared representers \(I=0,1,2\) across data sources to design varying levels of covariates shift, and introduce posterior drift by setting \(\beta^{(s)}_m = \beta_m + \delta^{(s)}\) with \(\beta_{m,\text{supp}}=(1,1)\) and \(\delta^{(s)} \sim \mathcal{N}(0,\sigma^2)\) for \(\sigma \in \{0.1, 1, 3\}\). For each setting, we perform 100 Monte Carlo replications. Instead of evaluating the predictive performance on the underlying non-missing data as in \citep{yu2020optimal}, the validation and testing sets are generated with the same missing pattern as specified in Figure \ref{fig: BR2_sim_missing_pattern}, and predictions are made based on observed modalities. The proposed method aims to improve the predictive performance for all data sources, regardless of missing patterns, while \citep{yu2020optimal} focuses on estimating the underlying homogeneous coefficients accurately.

Figure \ref{fig: sim2_results} shows the root mean squared error (RMSE) under various settings of $I, \sigma$ for $L=3$, where DISCOM is only applicable. The results for $L=1, 2$ can be found in Appendix \ref{app_sec: add_br2_sim}. As $\sigma$ increases, all methods perform worse, while our $BR^2$ learning performs consistently better than the baseline STL and DISCOM, especially in the highly heterogeneous scenarios (e.g., $I=0, \sigma=3$). In contrast, DISCOM can only outperform the baseline STL method when $I=2$ and $\sigma=0.1$, which is sensible because their homogeneous model assumption.

\section{Real Data Analysis}
\label{sec: real-data}

In this section, we apply our proposed method, $BR^2$ learning to ADNI data \citep{mueller2005alzheimer} to evaluate its performance. Existing methods compared in Section \ref{sec: simulation} have also been assessed in these examples. Additionally, we analyze the pan-cancer example where all the features are consistently observed in all data sources and $R^2$ learning applies. The detailed analysis and results can be found in Appendix \ref{app_sec: pan_cancer}.

Our analysis aims to learn an accurate predictor for cognitive status, measured by the Mini-Mental State Examination (MMSE, \citealt{folstein1975mini}). The covariates are from three modalities: magnetic resonance imaging (MRI), positron emission tomography (PET) and gene expression. The block missing structure emerges in the second phase of the ADNI study at the 48th month, where the detailed blockwise missing pattern is . In our analysis, we follow the pre-processing steps in \citep{xue2021integrating} which extract the region of interest (ROI) level data in ADNI and obtain $q_1 = 267$ features for MRI and $q_2 = 113$ feature for PET. For the genetic features, sure independence screening (SIS, \citealt{fan2008sure}) is adopted and $q_3 = 300$ genetic features are retained. In total, we use $p = q_1 + q_2 + q_3 = 680$ features to build our predictor, and four data sources with varying observation patterns are considered, where the first source with $n_1$ observe all three modalities, while the rest three sources have one different missing modality, respectively.

\begin{figure}[t]
    \centering
    \scalebox{0.5}{\begin{tikzpicture}
    
    \matrix[matrix of nodes, ampersand replacement=\&, nodes={anchor=center}] (data) at (0, 0) {
    \& {MRI $q_1 = 267$} \& {PET $q_2 = 113$} \& {Gene $q_3 = 300$} \\
    {$n_1=69$}\& |[draw, fill=lightblue1, minimum height=27.6mm, minimum width=53.4mm]|\& |[draw, fill=green3, minimum height=27.6mm, minimum width=22.6mm]| \& |[draw, fill=lightred2, minimum height=27.6mm, minimum width=60mm]| \\
    {$n_2=96$}\&|[draw, fill=lightblue1, minimum height=38.4mm, minimum width=53.4mm]|\& |[draw, fill=green3, minimum height=38.4mm, minimum width=22.6mm]| \& |[draw, fill=white, minimum height=38.4mm, minimum width=60mm]| \\
    {$n_3=34$}\&|[draw, fill=lightblue1, minimum height=13.6mm, minimum width=53.4mm]|\& |[draw, fill=white, minimum height=13.6mm, minimum width=22.6mm]| \& |[draw, fill=lightred2, minimum height=13.6mm, minimum width=60mm]| \\
    {$n_4=13$}\&|[draw, fill=white, minimum height=5.2mm, minimum width=53.4mm]|\& |[draw, fill=green3, minimum height=5.2mm, minimum width=22.6mm]| \& |[draw, fill=lightred2, minimum height=5.2mm, minimum width=60mm]| \\
    };

    \matrix[matrix of nodes, ampersand replacement=\&, nodes={anchor=center}] (response) at (9, 0) {
    {MMSE} \\
    |[draw, minimum width=10mm, minimum height=27.6mm, fill=lightpurple2]| \\
    |[draw, minimum width=10mm, minimum height=38.4mm, fill=lightpurple2]| \\
    |[draw, minimum width=10mm, minimum height=13.6mm, fill=lightpurple2]| \\
    |[draw, minimum width=10mm, minimum height=5.2mm, fill=lightpurple2]| \\
    };

\end{tikzpicture}}
    \caption{\label{fig: ADNI_structure} Blockwise missing pattern of ADNI data. White blocks are missing modalities. }
\end{figure}

In this analysis, we compare our method with Single Task Learning (STL), DISCOM \citep{yu2020optimal}, and mean-imputation followed by $R^2$ learning ($R^2$-Impute). To evaluate the performance of our proposed $BR^2$ learning and competing methods, we split the data into training, validation and test set randomly with proportion 60\%, 20\% and 20\%, respectively. Note that the sampling procedure follows stratified sampling, for example, the training set consists of 60\% samples from each data source. We measure the performance with root mean squared error (RMSE) and repeat the above sampling procedure 100 times. The results are shown in Table \ref{tab: adni_results}.

\begin{table}[H]
    \centering
    \begin{tabular}{lcccccc}
    \toprule
    Data Source & $BR^2$ learning & $R^2$-Impute &  STL & DISCOM\\
    \midrule
    1 & \textbf{3.687(1.186)} & 3.763(1.042) & 3.901(1.167) & 12.743(7.881) \\
    2 & 2.578(0.719) & \textbf{2.527(0.607)} & 2.620(0.655) &  27.082(1.755)\\
    3 & \textbf{5.039(1.340)} & 5.203(1.588) & 5.566(1.521) & 13.309(6.136)\\
    4 & \textbf{3.272(1.416)} & 3.708(1.252) & 3.828(1.014) & 10.936(7.149)\\
    \bottomrule
    \end{tabular}
    \caption{\small RMSEs of proposed $BR^2$ learning and competing methods on four data sources.}
    \label{tab: adni_results}
\end{table}

The proposed $BR^2$ learning achieves the minimal RMSE over three data sources, and clearly improves performance compared to STL and DISCOM consistently, since our $BR^2$ learning can effectively integrate available information for block-missing data. Given the exploratory analysis in Appendix \ref{app_sec: motivation}, covariates shift and posterior drift are non-negligible across the four data sources in this dataset, which deteriorates the performance of DISCOM. In contrast, our method is robust against these sources of heterogeneity and effectively extract shared information to achieve more accurate prediction. In addition, $R^2$-Impute also improves predicitive performance over STL across all data sources, and achieves the minimal RMSE in the second data source. Such results reflect the debate between imputation-based and imputation-free methods for handling missing data problems. As an imputation-free approach, $BR^2$ learning achieves significant improvement over DISCOM, however, it is still an open question that which type of methods should be adopted for a real data problem with few knowledge of missing mechanism.

\section{Discussion}
\label{sec: discussion}

This paper investigates data integration in supervised learning under three types of heterogeneity: covariate shift, posterior drift, and blockwise missingness. Accommodating these heterogeneity enables practitioners to integrate a larger number of data sources and improve predictive performance. The proposed \emph{Block-wise Representation Retrieval} learning method is, to the best of our knowledge, the first approach in the literature that simultaneously accommodate all these heterogeneity. Moreover, we introduce the notion of \emph{integrativeness} for representers and propose a novel \emph{Selective Integration Penalty} that promotes more integrative representers in the dictionary, thereby enhancing model generalization both empirically and theoretically. The effectiveness of our approach is demonstrated through extensive simulation studies and real data applications.

Several directions for future research emerge from our work. First, in the multi-modality block-missing problem, our method exploits all observed modalities to improve prediction accuracy, following the so-called imputation-free strategy. Alternative approaches, summarized as imputation-based strategy (e.g., \citealt{xue2021integrating, ma2025deep}) remains prevalent. Our numerical studies suggests there is no uniform winner between these two strategies, it is still unclear which strategy is favorable under specific conditions. Future work could explore quantitative criteria to distinguish the circumstances favoring one strategy over the other. Second, it is of great interest to develop predictive inference procedures under heterogeneous data. Although our method and existing methods show promising performance in prediction, it is still unclear how to borrow information across heterogeneous data to improve the uncertainty quantification, say, narrower predictive interval. Third, we consider a static data analysis paradigm that all data sources are collected and fixed beforehand. In real applications, it is common that data sources are collected in an online fashion. This problem is formulated as streaming data analysis \citep{luo2023multivariate, luo2023statistical} in statistics and continual learning \citep{kirkpatrick2017overcoming, van2019three, rusu2016progressive} in machine learning. It is of great interest to extend our $R^2$ or $BR^2$ learning to adapt online data integration.

\begin{spacing}{0.9}  
{\footnotesize
\bibliographystyle{apalike}
\bibliography{sample}
}
\end{spacing}
\end{document}